\documentclass{article}
\usepackage{spconf,amsmath,graphicx}
\usepackage{bm,amssymb,hyperref,diagbox}

\title{e-ACJ: Accurate Junction extraction for event cameras}
\name{Zhihao Liu, Yuqian Fu}
\address{Electronic Information School, Wuhan University, China}

\begin{document}
%
\maketitle
\begin{abstract}
  Junctions reflect the important geometrical structure information of the image, and are of primary significance to applications such as image matching and motion analysis. Previous event-based feature extraction methods are mainly focused on corners, which mainly find their locations, however, ignoring the geometrical structure information like orientations and scales of edges. This paper adapts the frame-based \emph{a-contrario} junction detector(ACJ) to event data, proposing the event-based \emph{a-contrario} junction detector(e-ACJ), which yields junctions' locations while giving the scales and orientations of their branches. The proposed method relies on an \emph{a-contrario} model and can operate on asynchronous events directly without generating synthesized event frames.  We evaluate the performance on public event datasets. The result shows our method successfully finds the orientations and scales of branches, while maintaining high accuracy in junction's location.
\end{abstract}
\section{INTRODUCTION}
\label{sec:intro}
Junctions are of primary importance for visual perception and scene understanding. They are usually classified into L-,Y-(or T-) and X-junctions. Junctions are usually used as important keys for various computer vision tasks, such as Simultaneous Localisation And Mapping (SLAM) \cite{slam}, image segmentation, figure/ground separation \cite{ren2006figure} \cite{dimiccoli2009exploiting}, and many others \cite{lin2007empirical} \cite{shen2000uncertainty}.

The detection of junction features in frame-based images has been studied for many years \cite{maire2008using} \cite{puspoki2015design}. The scaling of junctions for recognition purposes has also been studied \cite{mikolajczyk2004scale}. These methods determine the scale of the points of interest in a localized area, which significantly reduces the accuracy. In addition, these methods focus mainly on the localization and scale of corner points and ignore the differences between different types of junctions.
To overcome these shortcomings, \cite{xia2014accurate} proposed the ACJ detector to detect and characterize different junctions with non-linearly scaled points. In the ACJ detector, to determine the position and bifurcation of junctions, the \emph{a-contrario} are applied to eliminate the effect of noise and to obtain accurate results. 
On the other hand, the typical junction detection process in a conventional camera is not always optimal for robot tasks, because many images contain the same redundant information when they are not moving, resulting in a large waste of computational resources. Furthermore, robots are challenged to detect junctions in the dark.
To overcome the challenges of high speed, dynamic range, and latency, event cameras provide a wealth of information encoded in event streams for AR/VR and robotics. Unlike traditional frame-based cameras, event cameras respond to pixel-by-pixel brightness changes with high temporal resolution and transmit events asynchronously with very low latency. They also have a high dynamic range and are more resistant to motion blur. Applying the event camera as a detection junction to a robotic task also solves the problem of wasting computational resources, since the event camera does not generate new data for the same image (when there is no change in brightness).

We proposed a method combining event camera and frame-based ACJ detector ({\bf e-ACJ}). In the meantime, the detection method of ACJ is improved for the special nature of event camera, and Arc$^*$ algorithm \cite{alzugaray2018asynchronous} is introduced as well to accelerate the original ACJ detection speed. We also applies techniques such as \emph{a-contrario} and binarization to the event camera data. The results show that the accuracy of e-ACJ detection is better than most event camera corner detection algorithms, such as eHarris \cite{eharris}, eFAST \cite{efast} and FA-Harris \cite{li2019fa}.
In the following section, we address the ACJ detector algorithm based on event camera data  (e-ACJ). Section \ref{sec:algorithm} presents the details and implementation of the proposed e-ACJ detector. Section \ref{sec:experiment} gives the method for the collection of ground truth event-junctions, the evaluational methods and gives the experimental results. Finally, conclusions are drawn in section \ref{sec:conclusions}.
\section{THE ALGORITHM}
\label{sec:algorithm}
Due to the demand for an efficient event-based junction detector, we now propose a novel accurate and fast event-based junction detector, called e-ACJ. The coming event is firstly used to generate a global Surface of Active Events(G-SAE)\cite{li2019fa}. Afterwards, an event filter based on SAE is used to select event-junctions candidate, as described in section \ref{ssec:accelerate}, which will efficiently reduce time consumption. Then our proposed e-ACJ detector are employed on event-junctions candidates, getting rid of junctions that don't meet junction-ness needed. Finally a junction refinement is used to improve the results. The following sections elaborate on the proposed method.
\subsection{G-SAE Construction and Update}
The event cameras do not output intensity-frames like traditional cameras, but a stream of events including the positions and timestamps. An event is denoted by $(x,y,t,p)$, where $(x,y)$ is the position where the event occurs in image plane, $t$ is the timestamp and $p (\pm 1)$ is the polarity of this event. Since making use of all the previous events to find junctions is impossible, a widely used method in event-stream processing is SAE, defined as SAE$:(x,y) \in \mathbb{R}^2 \mapsto t\in \mathbb{R}$.

\cite{eharris} establishes a 9$\times$9 local SAE for every pixel in the image plane. It increases the processing time dramatically and makes SAEs maintenance harder, as multiple SAEs need to be refreshed as well as the number of SAEs needing to be maintained equals the resolution of the event camera. \cite{li2019fa} proposed global SAE(G-SAE), which is the same size to the image plane. For each pixel,G-SAE saves the latest event timestamp at the corresponding position. When a new event comes, G-SAE will update the corresponding timestamp at which this event occurs. Thus only one G-SAE is maintained, and local SAE centered on the event can be extract from G-SAE directly.
\subsection{e-ACJ Junction Detection}
\begin{figure}[htbp]
  \begin{minipage}[b]{\linewidth}
    \centering
    \centerline{\includegraphics[width=.5\linewidth]{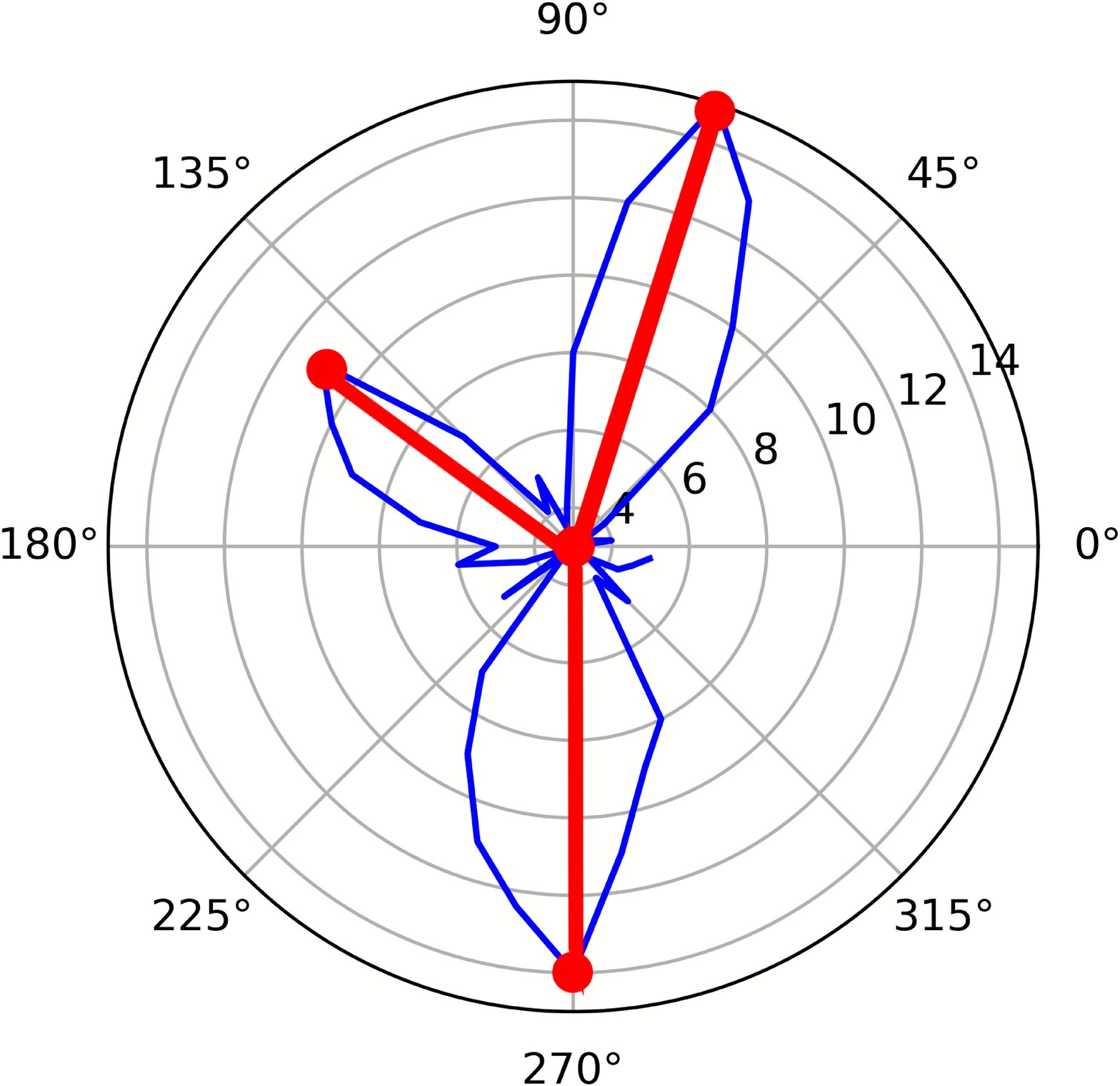}}
  \end{minipage}
  \caption{Branch strength in blue and semi-local maxima in red \cite{xia2014accurate}}
  \label{fig:branch}
\end{figure}

\cite{xia2014accurate} has proposed ACJ detection method for images of conventional cameras. An anisotropic scale junction with \textit{M} branches is denoted as
\begin{equation}
  \jmath=
  \begin{Bmatrix}
    \bm{p},\{{r_i,\theta_i}\}_{i=1}^M
  \end{Bmatrix}
\end{equation}
where $\bm{p}$ is the center location, $r_i$ is the scale for $i$-th branch and $\theta_i$ orientation for it. Scale of junction is characterized by it's minimal branch scale. Since junction can be seen as intersection of several lines from different direction, \cite{xia2014accurate} computed the junction-ness for a branch with a given scale and orientation, and then exploited \emph{a-controrio} method to judge whether branch is meaningful enough in input image. The junction-ness for given scale $r$ and $\theta_i$ is computed in a sector region, as shown in Fig.\ref{fig:branch}. The sector region for given location $\bm{p}$, scale $r$ and orientation $\theta$ can be mathematically described as
\begin{equation}
  \begin{aligned}
    S_p(r,\theta):=\{q\in\Omega ;q\neq p, & \left \|\vec{pq}\right \|\leq r,                   \\
                                          & d_{2\pi}(\alpha(\vec{pq}),\theta)\leq \Delta(r) \}
  \end{aligned}
\end{equation}
where $\Omega$ is the input image, $\alpha(\vec{pq})$ denotes the angle of vector $\vec{pq}$, $d_{2\pi}(\alpha,\beta)$ is described by function $d_{2\pi}(\alpha,\beta)=min(|\alpha-\beta|,2\pi-|\alpha-\beta|)$, and $\Delta r$ is defined as $\frac{\tau }{r}$ with a predefined parameter $\tau$ to control the width of sector region.
Since branches are lines where the brightness changes,the normal angle for gradient should share the direction with the branches' orientation. In this way, the more points $\bm{q}\in S_p(r,\theta)$ in a branch's in sector region have close normal angles with orientation $\theta$, the more likely it's a branch of corresponding junction with scale $r$ and $\theta$. Followed with this idea,for a given branch and it's sector region $S_p(r,\theta)$, the junction-ness of a branch can be measured by
\begin{equation}
  \omega_{\bm{p}}(r,\theta)=\sum_{\bm{q}\in S_{\bm{p}}(r,\theta)} \gamma_{\bm{p}}(\bm{q})
\end{equation}
$\gamma_{\bm{p}}(\bm{q})$ is the junction-ness of point $\bm{q}$ relative to position $\bm{p}$. For each coming event, we apply a local patch centered in this event with size $(2r+1)\times (2r+1)$, where $r$ ranges from 3 to 15 in our experiment. Then we make this region an essential binary patch contains only \texttt{0} and \texttt{1}. The latest $\left \lceil (r+1)^2 \right \rceil$ events are \texttt{1}, otherwise \texttt{0}. The meaning of gradient modulus in binarized images changed from how fast the brightness changes to whether the brightness changes. Therefore,we only take bool values \{\texttt{0},\texttt{1}\} for the norm value of gradient, which indicates whether the brightness changes at the given position. Thus, we can define
$\gamma_{\bm{p}}(\bm{q})$ as
\begin{equation}
  \begin{aligned}
    \gamma_{\bm{p}}(\bm{q})=\left \| \tilde{I}(\bm{q}) \right \|\cdot max(| & cos(\phi \bm(q)-\alpha(\vec{pq}))|      \\
                                                                            & -|sin(\phi \bm(q)-\alpha(\vec{pq}))|,0)
  \end{aligned}
\end{equation}
where the $\left \| \tilde{I}(\bm{q}) \right \|$ is the bool values representing the norm of gradient at $\bm{p}$, $\phi (\bm{q})$ is the orientation of normal angle at $\bm{q}$, defined as $\phi (\bm{q})=(arctan\frac{I_y(\bm{q})}{I_x(\bm{q})} +\frac{\pi}{2})~modulo~(2\pi)$. $I_x,I_y$ are the directional derivative in the $x$ and $y$ direction, which is obtained by applying a Sobel 3$\times$ 3 filter.

Junction-ness for an entire junction is described by it's minimal junction-ness branch as
\begin{equation}
  t(\jmath):=\min_{m=1,...,M}\omega_{\bm{p}}(r,\theta_m)
\end{equation}
where $M$ is the number of branches for junction $\jmath$.
With a given position $\bm{p}$ and orientation $\theta_i$, junction detection problem can be simplified to only estimating scale $r_i$. One feasible choice is \textit{a-contrario} method, which determines if junction is $\varepsilon$-meaningful with respect to given $r$ and orientation. The $\varepsilon$-meaning scale $r$ uses cumulative distribution function (CDF) and can be formulated to:
\begin{equation}
  F(t;J(r,\theta))=P\{\omega_{\bm{p}}\geq t\}=\int_{t}^{+\infty}d(  \mathop{\star}\limits_{j=1}^{J(r,\theta)}h)
\end{equation}
where $\gamma_{\bm{p}}(\bm{q})$ is the distribution of random variable $\omega_p(r,\theta)$, $J(r,\theta)$ is the number size of corresponding sector, and $\mathop{\star}\limits_{j=1}^{J(r,\theta)}h$ means convolutional result for $\gamma_{\bm{p}}(\bm{q})$ probability density function with $J(r,\theta)$ times. Under the null hypothesis in \cite{xia2014accurate}, $\left \| \tilde{I}(\bm{q}) \right \|$ and $\phi(\bm{q})$ are made up of independent random variables and $\phi\bm{q}$ is uniformly distributed in $[0,2\pi]$.
As $\alpha(\vec{pq})$ is a constant with given position $\bm{p}$, the angle $\phi(\bm{q})-\alpha(\vec{pq})$ is still uniformly distributed in $[0,2\pi]$. Since $\left \| \tilde{I}(\bm{q}) \right \|$
only take values between \texttt{0,1}, it can be described as a two-point distribution $\left \| \tilde{I}(\bm{q}) \right \|\sim B(1,p)$.Finally the distribution of $\gamma_{\bm{p}}(\bm{q})$ can derived as
\begin{equation}
  h(z)=\left\{\begin{array}{l}
    (1-\frac{p}{2})\delta(z)+\frac{2p}{\pi\sqrt{2-z^2}},0\leq z\leq 1 \\
    0,otherwise
  \end{array}\right.
\end{equation}
$p$ is computed from real event datasets. We have tested on six datasets, with movement and texture from simple to complex, for six million events in total. For each event tested, we extract the local patch centered it and apply a Sobel 3$\times$ 3 filter in local patch, then compute the $p$ for this event. We calculate the average $p$ value for each dataset, and the biggest $p$ value is chosen as our parameter, since a bigger $p$ will guarantee less mistakes. As a local patch is at most $(2r+1)\times (2r+1)$ size, six million can describe the possible situation well. The biggest $p$ in our test is 0.21 and we select it as our $p$ parameter. Thus the $\varepsilon$-meaningful scale can be determined with \textit{a-contrario} approach
\begin{equation}
  {\rm NFA}(\jmath):=\#\mathcal{J}(M)\cdot F(t;J(r,\theta))\leq \varepsilon
\end{equation}
where $\#\mathcal{J}(M)$ is the number of test for junctions with $M$ branches. A junction is $\varepsilon$-meaning if its average number of false alarm (NFA) is smaller than $\varepsilon$. In our experiment $\varepsilon$ is set to \texttt{1}, meaning that no false alarms are expected.
\subsection{Detection Refinement}
In \cite{xia2014accurate} a refinement operation is applied on junctions which have distances smaller than a predefined parameter $r_d$ between each other, and the one with smaller NFA is removed.This can also be applied on e-ACJ method, just notice that we need focus on not only their positions but also timestamps to avoid error removal from the time stream.We set parameter $T$ to indicate maximum time interval between two junctions. If the absolute value of the difference between their timestamps is greater than $T$, then either of them is removed, since we consider they happen at difference time. $T$ is set to $0.005s$ in our experiment.
\subsection{Speed Up}
\label{ssec:accelerate}
ACJ for frame cameras make use of Line Segment Detector(LSD) \cite{lsd} to select potential junctions. However, it's impossible to employ LSD on e-ACJ since we detect junctions at the time when events come and never generate frames. In order to enhance the real-time performance of e-ACJ, we use the Arc$^*$ approach proposed by \cite{alzugaray2018asynchronous}, which runs faster but detects more false corners as well. For coming event-stream, we first apply Arc$^*$ to filter the events and select junction candidates. Time-consuming computations and refinement will only be used on selected junction candidates. With an Arc$^*$ filter our e-ACJ detector achieves about 10$\times$ speed-up compared to the original method without accelerating.
\section{EXPERIMENTS}
\label{sec:experiment}
We evaluated the proposed junction detection algorithm by running it on Event Camera Datasets \cite{efast}. We carefully selected a representative subset with increasing complexity and event-rate (\texttt{shapes}, \texttt{dynamic}, \texttt{poster} and \texttt{boxes}) to ensure a comprehensive and fair evaluation scheme. The dataset was recorded by DAVIS-240C \cite{brandli2014240}, which contains many sequences of frame-based, such as intensity images and asynchronous events at the resolution of 240$\times$180. We perform our evaluation work on a laptop equipped with an Intel i7-8750H CPU with 2.20GHz and RAM with 8GB.

In the experiment, we get our ground-truth first and then followed the qualitative analysis of our e-ACJ. After that, we give the accuracy of our e-ACJ. The implementations of Arc$^*$ is provided by \cite{alzugaray2018asynchronous}. All methods are implemented with C++.
\subsection{Ground-Truth for Event Cameras}
\label{ssec:ground_truth}
Generating ground-truth for implementing event-based algorithms is particularly challenging due to the characteristics of the event stream. In this experiment, we combine the (original) Harris detector with the KLT tracker \cite{klt} and apply it to the available intensity frames in the dataset to track the intensity bifurcation points, similar to the method presented in \cite{alzugaray2018asynchronous}. For event corners that fall within a 3.5 pixel cylinder, we consider them to be true event junctions and label them as ground-truth. Also, for event junctions detected between 3.5 and 5 pixels, we label it as a false event junction. 
The above method is used to find the true-false event junction, which is used for the following evaluation.

\subsection{Qualitative Evaluation}
\label{ssec:qualitative_evaluation}
\begin{figure}[htbp]
  \begin{minipage}[b]{.48\linewidth}
    \centering
    \centerline{\includegraphics[width=4.0cm]{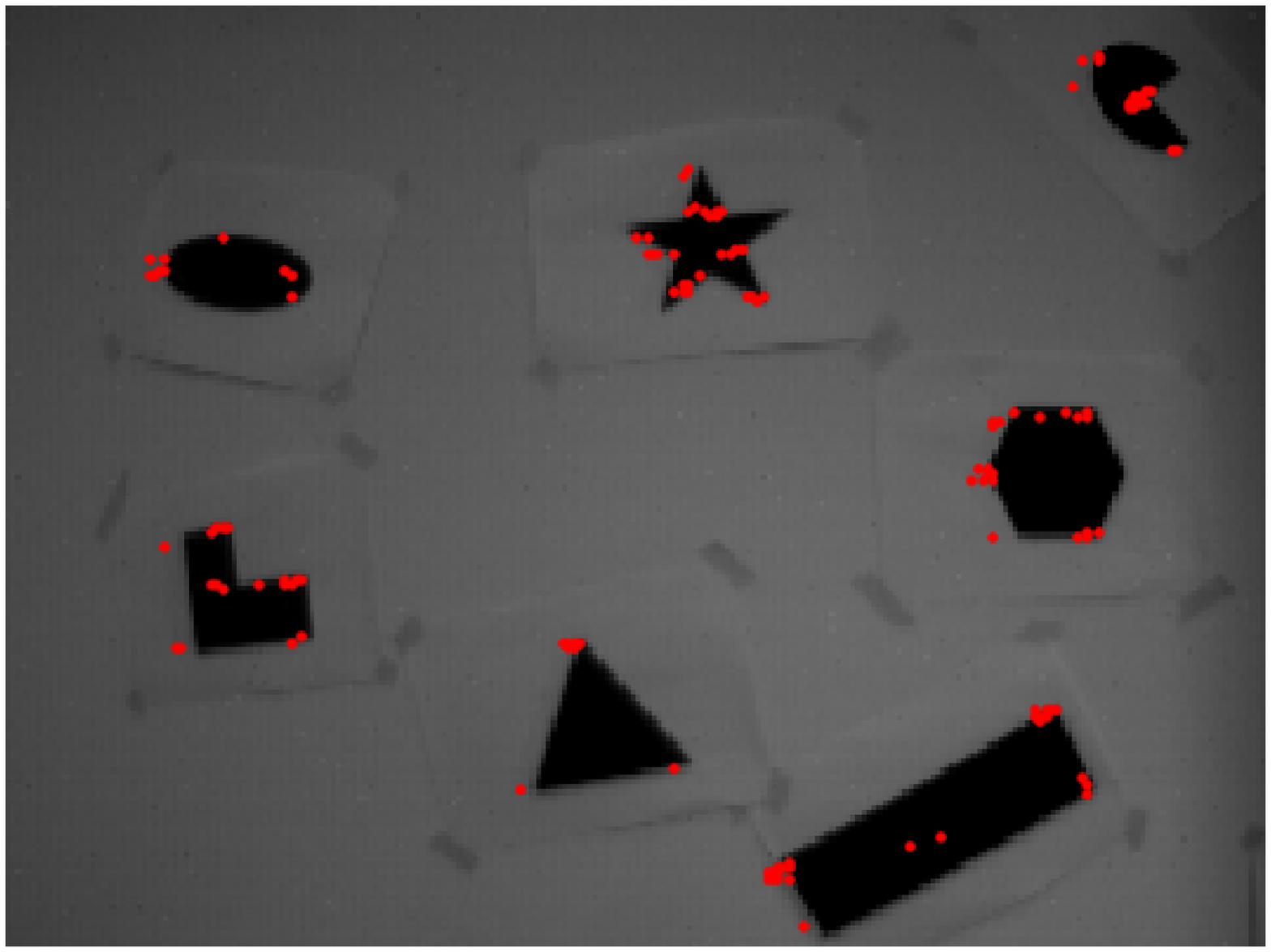}}
    \centerline{(a) shapes}\medskip
  \end{minipage}
  \begin{minipage}[b]{.48\linewidth}
    \centering
    \centerline{\includegraphics[width=4.0cm]{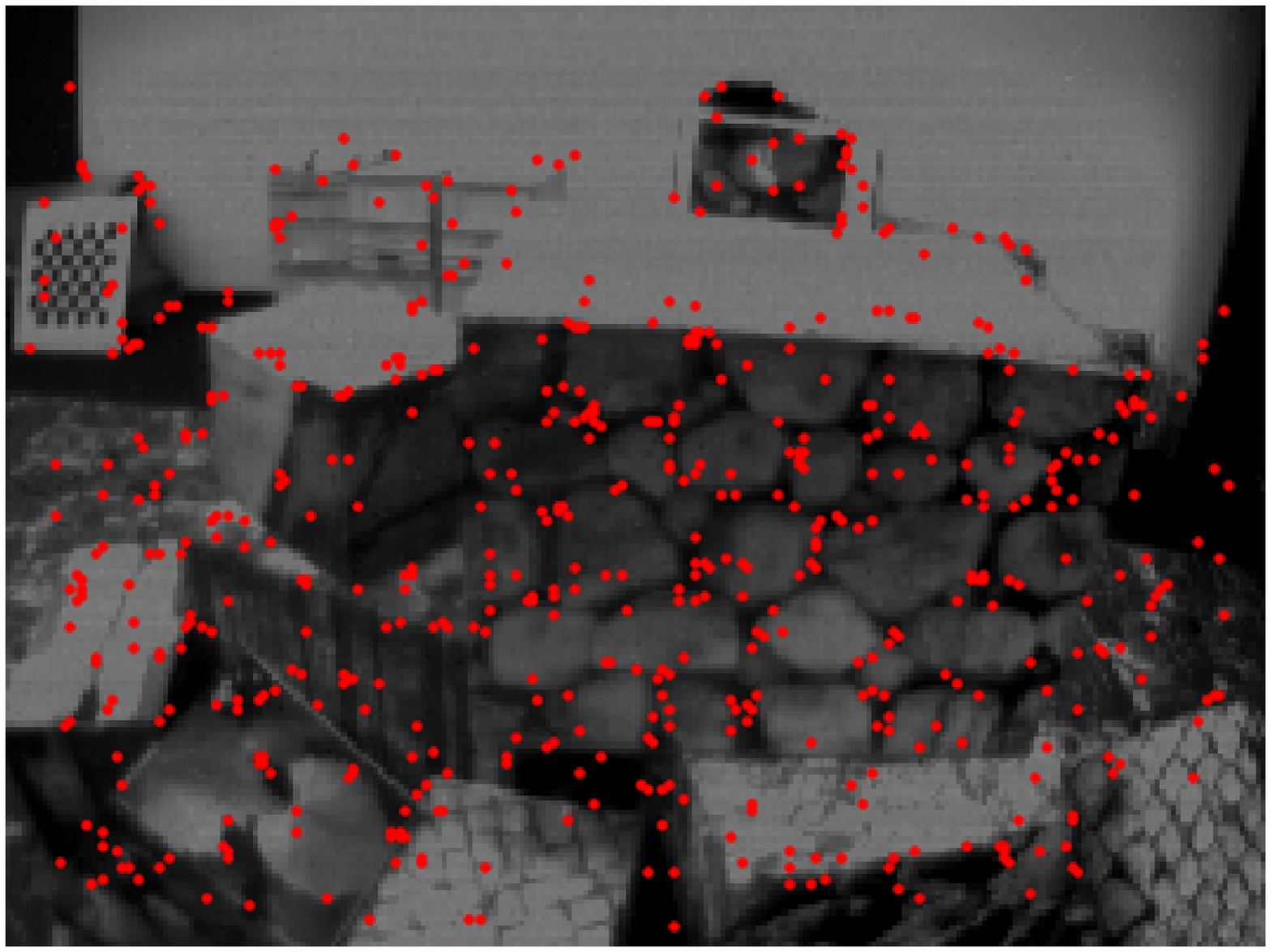}}
    \centerline{(b) boxes}\medskip
  \end{minipage}

  \begin{minipage}[b]{0.48\linewidth}
    \centering
    \centerline{\includegraphics[width=4.0cm]{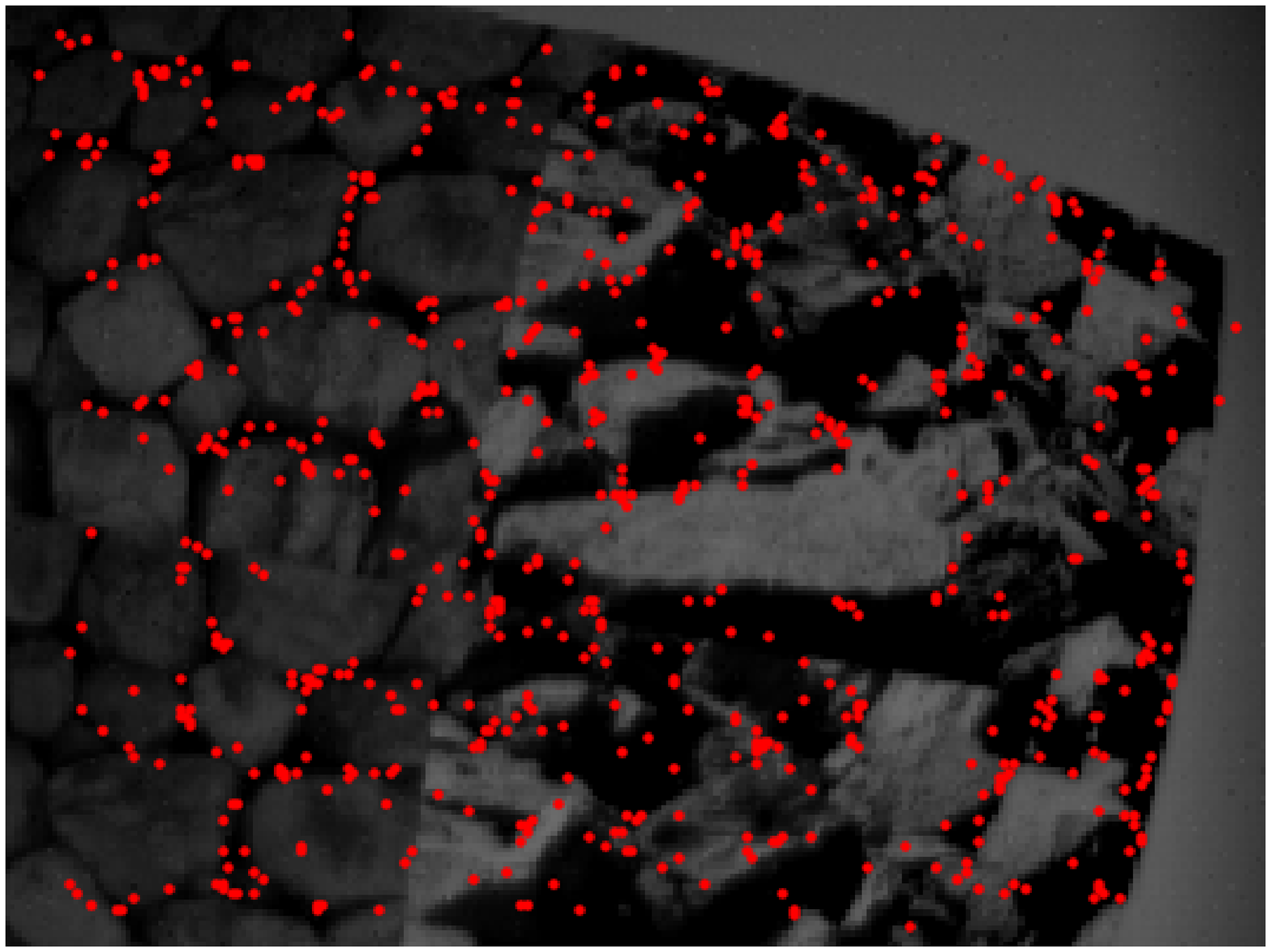}}
    \centerline{(c) poster}\medskip
  \end{minipage}
  \begin{minipage}[b]{.48\linewidth}
    \centering
    \centerline{\includegraphics[width=4.0cm]{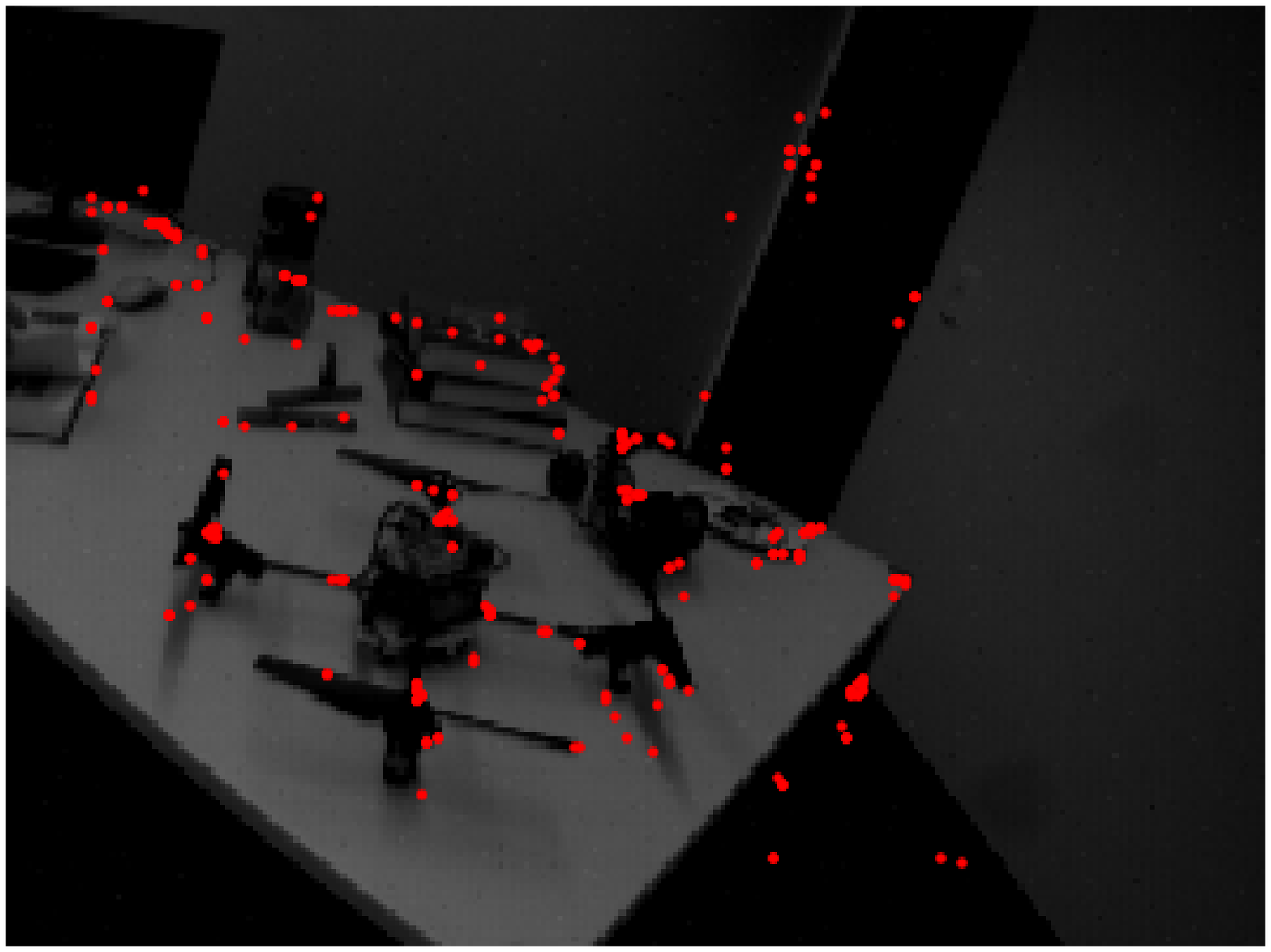}}
    \centerline{(d) dynamic}\medskip
  \end{minipage}
  \caption{The event-junctions (blue) detected by the e-ACJ detector within 50ms. The intensity-images are used for visualization. }
  \label{fig:qua-analysis}
\end{figure}

\begin{figure}[htbp]
  \begin{minipage}[b]{\linewidth}
    \centering
    \centerline{\includegraphics[width=.5\linewidth]{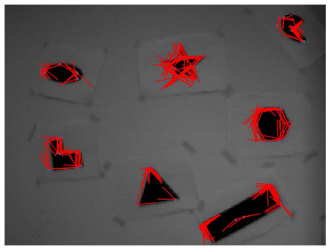}}
  \end{minipage}
  \caption{The junction branches with scale, and the scale of a branch is a measure of how well the corresponding angular sector agrees with the pixels it contains \cite{xia2014accurate}.}
  \label{fig:qua-analysis1}
\end{figure}

Fig.\ref{fig:qua-analysis} shows a qualitative evaluation of event-junctions detected by the e-ACJ detector for different datasets. We synthesize events within 50ms on the intensity-images for visualization. Fig.\ref{fig:qua-analysis} (a) indicates that our proposed e-ACJ detector can detect junctions with different angles. In Fig.\ref{fig:qua-analysis1} , we show the junction branches with scale, 
and the branch scale is a measure of how well the corresponding angular sector matches the pixel it contains.\cite{xia2014accurate}. The results shows that e-ACJ can obtain the scale information of  junctions well, which is convenient for subsequent feature tracking and other tasks.
\subsection{Accuracy}
\label{ssec:accuracy_performance}
Similarly to \cite{alzugaray2018asynchronous} and \cite{li2019fa}, 
we calculate the false positive rate and the accuracy of each method. The results are summarized in the table. Note that only the first 10 seconds of all scenes are used in this section. In \texttt{FPR}, we label an event as true positive (\texttt{TP}) if it is within 3.5 pixels of a tracked intensity junction, and false positive (\texttt{FP}) if it is between 3.5 and 5 pixels. Events that are not event junctions are then labeled as false negative (\texttt{FN}) within 3.5 pixels; events within the range of 3.5 to 5 pixels are labeled as true negative (\texttt{FP}). Then, \texttt{FPR} is calculated by \texttt{FPR=FP/(FP+TN)}. 

The accuracy indicates the percentage of correct event junctions among the detected event junctions. The results for accuracy evaluation is summarized in Table \ref{tab:accuracy}. As shown in Table \ref{tab:accuracy}, our e-ACJ can detect much more correct event-junctions and demonstrate better performance in terms of accuracy. However, since corners are not necessarily junctions (corners may not have {\bf intersections}), the accuracy of e-ACJ is lost on the Harris-based (corners) ground-truth.
The False Positive Rate of different event-corners detection
methods is shown in Table \ref{tab:fpr}. The results show that our e-ACJ has lower \texttt{FPR} than other methods.
\begin{table}[htbp]
  \centering
  \caption{The False Positive Rate (\%) of different event-corners detection methods on different scenes. The best results are made in bold.  \cite{li2019fa}}
  \label{tab:fpr}
  \resizebox{0.48\textwidth}{!}{%
  \begin{tabular}{c|cccc}
  \hline
  \textbf{\diagbox{Alg.}{Scene}}                   & \textbf{shapes} & \textbf{dynamic} & \textbf{poster} & \textbf{boxes} \\ \hline
  \textbf{eHarris\cite{eharris}}     & 13.63  & 8.15    & 3.04   & 7.42        \\
  \textbf{eFAST\cite{efast}}      & 18.62  & 4.59    & 3.04   & 2.67         \\
  \textbf{Arc$^*$\cite{alzugaray2018asynchronous}}        & 29.88  & 18.10   & 13.34  & 11.13       \\
  \textbf{FA-Harris\cite{li2019fa}}  & 11.68  & 3.28    & 1.90   & 1.61         \\
  \textbf{e-ACJ}      & \textbf{1.10}   & \textbf{0.89}    & \textbf{1.54}   & \textbf{1.02}         \\ \hline
  \end{tabular}%
  }
  \end{table}
\begin{table}[htbp]
  \centering
  \caption{The accuracy (\%) of different event-junctions detection methods on different scenes. The best results are made in bold. \cite{li2019fa}}
  \label{tab:accuracy}
  \resizebox{0.48\textwidth}{!}{
  \begin{tabular}{c|cccc}
    \hline
    \textbf{\diagbox{Alg.}{Scene}}                   & \textbf{shapes} & \textbf{dynamic} & \textbf{poster} & \textbf{boxes}  \\ \hline
    \textbf{eHarris\cite{eharris}}                   & 56.97           & 54.50            & \textbf{49.04}           & 49.26                            \\
    \textbf{eFAST\cite{efast}}                     & 56.22           & 54.86            & 48.30           & 48.60                            \\
    \textbf{Arc$^*$\cite{alzugaray2018asynchronous}} & 55.42           & 53.50            & 48.81           & 48.94                           \\
    \textbf{FA-Harris\cite{li2019fa}}                & 57.66           & \textbf{55.86}            & 48.91           & 49.66                           \\
    \textbf{e-ACJ}                                   & \textbf{72.84}           & 52.48           & 48.37           & \textbf{56.40}                               \\\hline
  \end{tabular}
  }
\end{table}
\section{CONCLUSIONS}
\label{sec:conclusions}
We present an accurate junction-detection algorithm (e-ACJ) that works on the asynchronous output data stream of event cameras and makes use of its low latency and asynchronous characteristics well. Meanwhile, due to the characteristic of the ACJ method, we can extract the scale features of the  junctions, so that we can better process the subsequent feature tracking tasks. The evaluation indicates that e-ACJ has high accuracy and low \texttt{FPR}. 
At the same time, by introducing the Arc$^*$ method, the original ACJ algorithm achieves about 10$\times$ speed-up. 

In the future, our work will include improving the detection speed further and develop the accuracy of event feature tracking.



\bibliographystyle{IEEEbib}
\bibliography{e_acj}

\end{document}